# Model Internals-based Answer Attribution for Trustworthy Retrieval-Augmented Generation

**Jirui Qi**[1*]  **Gabriele Sarti**[1*]  **Raquel Fernández**[2]  **Arianna Bisazza**[1]

[1]Center for Language and Cognition (CLCG), University of Groningen
[2]Institute for Logic, Language and Computation (ILLC), University of Amsterdam
{j.qi, g.sarti, a.bisazza}@rug.nl, raquel.fernandez@uva.nl

## Abstract

Ensuring the verifiability of model answers is a fundamental challenge for retrieval-augmented generation (RAG) in the question answering (QA) domain. Recently, self-citation prompting was proposed to make large language models (LLMs) generate citations to supporting documents along with their answers. However, self-citing LLMs often struggle to match the required format, refer to non-existent sources, and fail to faithfully reflect LLMs' context usage throughout the generation. In this work, we present MIRAGE – Model Internals-based RAG Explanations – a plug-and-play approach using model internals for faithful answer attribution in RAG applications. MIRAGE detects context-sensitive answer tokens and pairs them with retrieved documents contributing to their prediction via saliency methods. We evaluate our proposed approach on a multilingual extractive QA dataset, finding high agreement with human answer attribution. On open-ended QA, MIRAGE achieves citation quality and efficiency comparable to self-citation while also allowing for a finer-grained control of attribution parameters. Our qualitative evaluation highlights the faithfulness of MIRAGE's attributions and underscores the promising application of model internals for RAG answer attribution.[1]

## 1 Introduction

Retrieval-augmented generation (RAG) with large language models (LLMs) has become the de-facto standard methodology for Question Answering (QA) in both academic (Lewis et al., 2020b; Izacard et al., 2022) and industrial settings (Dao and Le, 2023; Ma et al., 2024). This approach was shown to be effective at mitigating hallucinations and producing factually accurate answers (Petroni et al., 2020; Lewis et al., 2020a; Borgeaud et al., 2022;

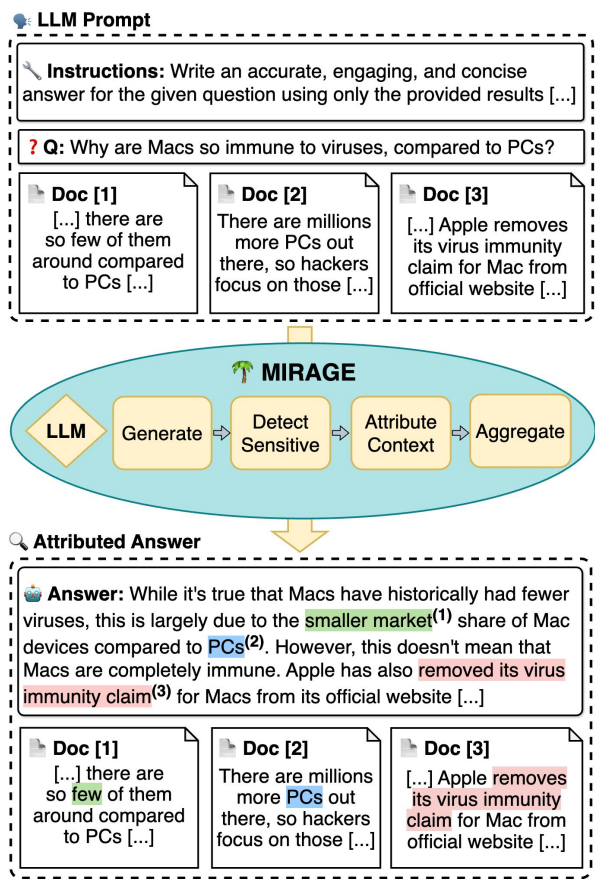

Figure 1: MIRAGE is a model internals-based answer attribution framework for RAG settings. Context-sensitive answer spans (in color) are detected and matched with contextual cues in retrieved sources to evaluate the trustworthiness of models' answers.

Ren et al., 2023). However, verifying whether the model answer is faithfully supported by the retrieved sources is often non-trivial due to the large context size and the variety of potentially correct answers (Krishna et al., 2021; Xu et al., 2023). In light of this issue, several *answer attribution*[2] approaches were recently proposed to ensure the trustworthiness of RAG outputs (Rashkin et al., 2021; Bohnet et al., 2022; Muller et al., 2023). Initial efforts in this area employed models trained on Natural Language Inference (NLI) to automate the identification of supporting documents (Bohnet et al., 2022; Yue et al., 2023). However, being based on an external validator, this approach does not faithfully explain the answer generation process but simply identifies plausible sources supporting model answers in a post-hoc fashion. Following recent progress in the instruction-following abilities of LLMs, *self-citation* (i.e. prompting LLMs to generate inline citations alongside their answers) has been proposed to mitigate the training and inference costs of external validator modules (Gao et al., 2023a). However, self-citation performance is hindered by the imperfect instruction-following capacity of modern LLMs (Mu et al., 2023; Liu et al., 2023), and resulting attributions are still predicted in an unintelligible, post-hoc fashion. This is an important limitation for these approaches, since the primary goal of answer attribution should be to ensure that the LLM is not 'right for the wrong reasons' (McCoy et al., 2019).

---

[*]Equal contribution.
[1]Code and data released at https://github.com/Betswish/MIRAGE.

[2]We use the term *answer attribution* (AA) when referring to the task of citing relevant sources to distinguish it from *feature attribution* methods used in MIRAGE.

In light of these considerations, in this work we introduce MIRAGE, an extension of the context-reliance evaluation PECORE framework (Sarti et al., 2024) that employs model internals for efficient and faithful answer attributions. MIRAGE first identifies context-sensitive tokens in a generated sentence by measuring the shift in LM predictive distribution caused by the added input context. Then, it attributes this shift to specific influential tokens in the context using gradient-based saliency or other feature attribution techniques (Madsen et al., 2022). We adapt this approach to the RAG setup by matching context-dependent generated sentences to retrieved documents that contribute to their prediction and converting the resulting pairs to citations using the standard answer attribution (AA) format. We begin our assessment of MIRAGE on the short-form XOR-AttriQA dataset (Muller et al., 2023), showing high agreement between MIRAGE results and human annotations across several languages. We then test our method on the open-ended ELI5 dataset (Fan et al., 2019), achieving AA quality comparable to or better than self-citation while ensuring a higher degree of control over attribution parameters. In summary, we make the following contributions:

- We introduce MIRAGE, a model internals-based answer attribution framework optimized for RAG applications.

- We quantify the plausibility of MIRAGE attributions on two datasets, showing improvements over NLI and self-citation methods while ensuring better controllability and efficiency.

- We analyze challenging attribution settings, highlighting MIRAGE's faithfulness to LLMs' reasoning process.

## 2 Background and Related Work

In RAG settings, a set of documents relevant to a user query is retrieved from an external dataset and infilled into an LLM prompt to improve the generation process (Petroni et al., 2020; Lewis et al., 2020a). *Answer attribution* (Rashkin et al., 2021; Bohnet et al., 2022; Muller et al., 2023) aims to identify which retrieved documents support the generated answer (*answer faithfulness*, Gao et al., 2023b), e.g., by exploiting the similarity between model outputs and references.[3] Simplifying access to relevant sources via answer attribution is a fundamental step towards ensuring RAG trustworthiness in customer-facing scenarios (Liu et al., 2023).

### 2.1 Answer Attribution Methods

**Entailment-based Answer Attribution** Bohnet et al. (2022) and Muller et al. (2023) propose to approximate human AA annotations with NLI systems such as TRUE (Honovich et al., 2022), using a source document as premise and an LLM-generated sentence as entailment hypothesis. AAs produced by these systems were shown to correlate strongly with human annotations, prompting their adoption in AA studies (Muller et al., 2023; Gao et al., 2023a). Despite their effectiveness, entailment-based methods can be computationally expensive when several answer sentence-document pairs are present. Moreover, this setup assumes the NLI model ability to robustly detect entailment relations across all domains and languages for which the LLM generator is used. In practice, however, NLI systems were shown to be brittle in challenging scenarios, exploiting shallow heuristics (McCoy et al., 2019; Nie et al., 2020; Sinha et al., 2021; Luo et al., 2022), and require dedicated efforts for less-resourced settings (Conneau et al., 2018).

---
[3]Popular frameworks such as LangChain (Chase, 2022) and LlamaIndex (Liu, 2022) support similarity-based citations using vector databases.

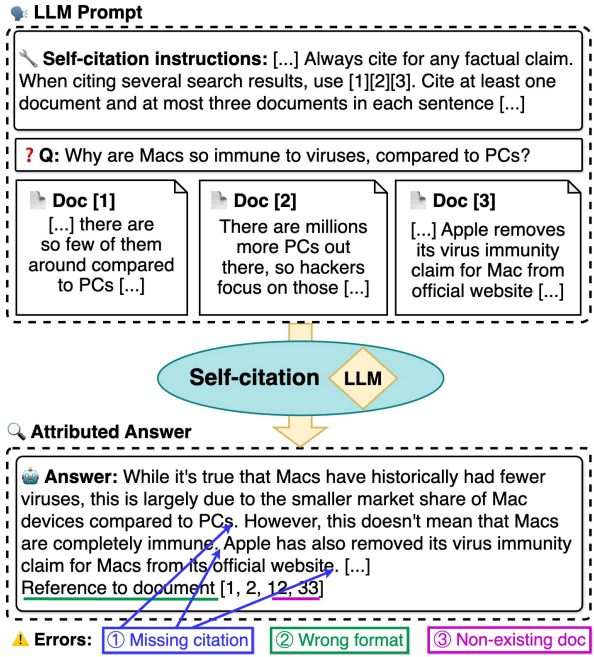

Figure 2: Instruction-following errors in a *self-citation* example, using the setup of Gao et al. (2023a).

| Model | Missing citation (%) | |
|---|---|---|
| | Sentence | Answer |
| Zephyr 7B $\beta$ | 54.5 | 95.7 |
| LLaMA 2 7B Chat | 62.4 | 99.3 |

Table 1: % of unattributed sentences and answers with $\geq 1$ unattributed sentences on ELI5.

## 2.2 Attribution Faithfulness

**Answer Attribution can be Unfaithful** The aforementioned approaches do not account for attributions' *faithfulness*, i.e. whether the selected documents influence the LLM during the generation. Indeed, the presence of an entailment relation or high semantic similarity does not imply that the retrieved document had an influence on the answer generation process. This can be true in cases where LLMs may rely on memorized knowledge while ignoring relevant, albeit unnecessary, contextual information. Even in the case of self-citation, recent work showed that, while the justifications of self-explaining LLMs appear plausible, they generally do not align with their internal reasoning process (Atanasova et al., 2023; Madsen et al., 2024; Agarwal et al., 2024; Randl et al., 2024), with little to no predictive efficacy (Huang et al., 2023). By contrast, approaches based on model internals are designed to faithfully reflect input importance in motivating model predictions. For instance, Alghisi et al. (2024) explore the use of gradients-based attribution to locate salient history segments for various dialogical tasks.

Concurrent to our work, Phukan et al. (2024) and Cohen-Wang et al. (2024) have proposed other internals-based methods for granular AA of LLM generations. While the two-step approaches proposed in both works are similar to MIRAGE, they also differ in substantial ways. Notably, Phukan et al. (2024) derive attributions from embedding similarity, which does not capture the functional influence of context usage during the generation process. CONTEXTCITE (Cohen-Wang et al., 2024) instead fits a linear surrogate model to estimate the impact of ablating context segments over downstream answer probabilities. While this procedure approximates causal context influence, it still requires a sufficiently large context and many LLM forward passes to learn the surrogate model[4], ultimately providing a coarser attribution for the full generated output. On the contrary, MIRAGE efficiently estimates generated tokens requiring attribu-

For example, NLI may fail to correctly attribute answers in multi-hop QA settings when considering individual documents as premises (Yang et al., 2018; Welbl et al., 2018).

**Self-citation** Gao et al. (2023a) is a recent AA approach exploiting the ability of recent LLMs to follow instructions in natural language (Raffel et al., 2020; Chung et al., 2022; Sanh et al., 2022; Brown et al., 2020), thereby avoiding the need for an external validator. Nakano et al. (2021) and Menick et al. (2022) propose citation fine-tuning for LLMs, while Gao et al. (2023a) instruct general-purpose LLMs to produce inline citations in a few-shot setting. Self-citation answers are generally more relevant to the provided sources' contents, but can still contain unsupported statements and inaccurate citations (Liu et al., 2023). In our preliminary analysis, we find that self-citation often misses relevant citations, uses wrong formats, or refers to non-existing documents (Figure 2). For the ELI5 dataset (Fan et al., 2019), we find that LLaMA 2 7B Chat (Touvron et al., 2023) and Zephyr $\beta$ 7B (Tunstall et al., 2023) fail to produce AAs matching the prompt instructions for the majority of generated sentences, with almost all answers having at least one unattributed sentence when the Gao et al. (2023a) self-citation setup is used (Table 1).

---
[4]Authors suggest a minimum of 32 different ablations.

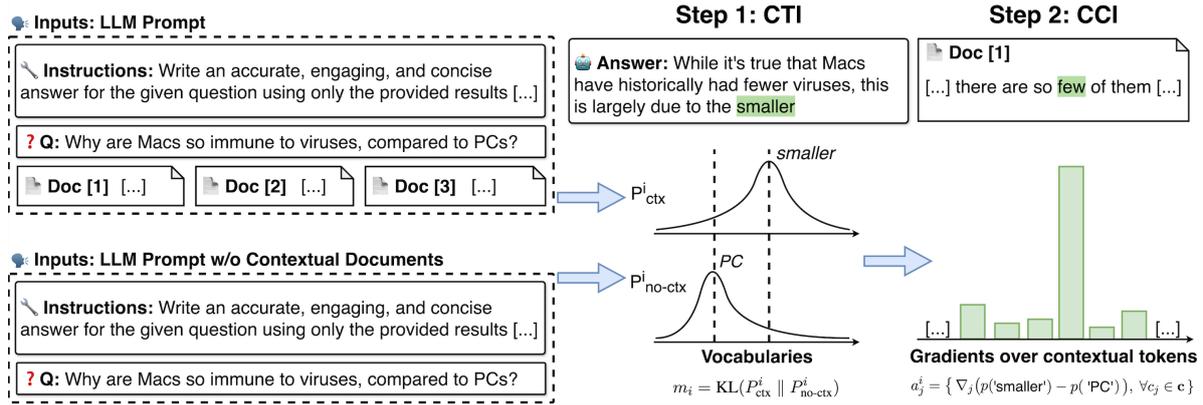

Figure 3: Illustration of MIRAGE's two-step approach adapted from PECORE for RAG answer attribution. **Step 1:** CTI detects context-sensitive tokens in the generation (e.g. *smaller*). **Step 2:** CCI attributes the generation of detected tokens back to context tokens (e.g. *few* in Doc[1] promotes the generation of *smaller* instead of *PC*) using contrastive feature attribution. Token pairs are then aggregated into sentence-document citations for practical usage.

tion via contrastive metrics to produce granular attributions at the token level, limiting computations to estimate how context impacts LLM predictions.

**Feature Attribution in Interpretability** The task of faithfully identifying salient context information has been studied extensively in the NLP interpretability field (Ferrando et al., 2024). In particular, *post-hoc feature attribution* approaches (Madsen et al., 2022) exploit information sourced from model internals, e.g., attention weights or gradients of next-word probabilities, to identify input tokens playing an important role towards the model's prediction. While feature attribution studies in NLP typically focused on classification tasks (Atanasova et al., 2020; Wallace et al., 2020; Chrysostomou and Aletras, 2022), recent work applies these methods to evaluate context usage in language generation (Yin and Neubig, 2022; Ferrando et al., 2023; Sarti et al., 2023, 2024). Importantly, feature attribution techniques are designed to maximize the faithfulness of selected context tokens by accessing models' intermediate computations, as opposed to the AA methods of Section 2.1. While the faithfulness of such approaches can still vary depending on models and tasks, the development of robust and faithful methods is an active area of research (Jacovi and Goldberg, 2020; Chan et al., 2022; Bastings et al., 2022; Lyu et al., 2024).[5]

---

[5] A maximally faithful AA approach would ablate all possible combinations of context elements to counterfactually estimate their importance towards model predictions. Given the long-form answers and contexts in RAG settings, this is practically unfeasible. Even if based on approximations, internals-based approaches such as MIRAGE are intrinsically more faithful than external validators like NLI, since they aim to exploit information functional to the predictive process

## 3 Method

Identifying which generated spans were most influenced by preceding information is a key challenge for LM attribution. The Model Internals-based RAG Explanations (MIRAGE) method we propose is an extension of the Plausibility Evaluation for Context Reliance (PECORE) framework (Sarti et al., 2024) for context-aware machine translation. Importantly, this framework requires open-weights access to the LLM generator, which is a strict but necessary requirement to provide an accurate overview of the actual context usage during generation (Casper et al., 2024). This section provides an overview of PECORE's two-step procedure (illustrated in Figure 3) and clarifies how MIRAGE adapts it for RAG answer attribution.

### 3.1 Step 1: Context-sensitive Token Identification (CTI)

For every token in an answer sentence $\mathbf{y} = \langle y_1, \ldots, y_n \rangle$ generated by a LM prompted with a query $\mathbf{q}$ and a context $\mathbf{c} = \langle c_1, \ldots, c_{|\mathbf{c}|} \rangle$, a contrastive metric $m$ such as KL divergence (Kullback and Leibler, 1951) is used to quantify the shift in the LM predictive distribution at the $i$-th generation step when the context is present or absent ($P^i_{\text{ctx}}$ or $P^i_{\text{no-ctx}}$). Resulting scores $\mathbf{m} = \langle m_1, \ldots, m_n \rangle$ reflect the context sensitivity of every generated token and can be converted into binary labels using a selector function $s_{\text{CTI}}$:

$$\text{CTI}(\mathbf{q}, \mathbf{c}, \mathbf{y}) = \{ y_i \mid s_{\text{CTI}}(m_i) = 1 \, \forall y_i \in \mathbf{y} \} \\ \text{with } m_i = \text{KL}(P^i_{\text{ctx}} \parallel P^i_{\text{no-ctx}}) \quad (1)$$

---

rather than only relying on the generated output.

## 3.2 Step 2: Contextual Cues Imputation (CCI)

For every context-sensitive token $y_i$ identified by CTI, a contrastive alternative $y_i^{\setminus \mathbf{c}}$ is produced by excluding $\mathbf{c}$ from the prompt, but using the original generated prefix $\mathbf{y}_{<i}$. Then, *contrastive feature attribution* (Yin and Neubig, 2022) is used to obtain attribution scores $\mathbf{a}^i = \langle a_1^i, \ldots, a_{|\mathbf{c}|}^i \rangle$ for every context token $c_j \in \mathbf{c}$:

$$\mathbf{a}^i = \big\{ \nabla_j \big( p(y_i) - p(y_i^{\setminus \mathbf{c}}) \big), \ \forall c_j \in \mathbf{c} \big\} \quad (2)$$

where $\nabla_j$ is the L2 norm of the gradient vector over the input embedding of context token $c_j$, and both probabilities are computed from the same contextual inputs $(\mathbf{q}, \mathbf{c}, \mathbf{y}_{<i})$. Intuitively, this procedure identifies which tokens in $\mathbf{c}$ influence the prediction of $y_i$ while accounting for the non-contextual option $y_i^{\setminus \mathbf{c}}$. Resulting scores are once again binarized with a selector $s_{\text{CCI}}$:

$$\text{CCI}(y_i) = \{ c_j \mid s_{\text{CCI}}(a_j^i) = 1, \ \forall c_j \in \mathbf{c} \} \quad (3)$$

This results in pairs of context-sensitive generated tokens and the respective input-context tokens influencing their prediction:

$$\mathcal{P} = \big\{ \langle y_i, c_j \rangle, \ \forall y_i \in \text{CTI}, \forall c_j \in \text{CCI}(y_i) \big\} \quad (4)$$

## 3.3 From Granular Attributions to Document-level Citations

**CTI Filtering** To obtain discrete labels from the CTI step, we set $s_{\text{CTI}}(m_i) = m_i \geq m^*$, where $m^*$ is a threshold value for selecting context-sensitive generated tokens. In this work, we experiment with two variants of $m^*$: a **calibrated threshold** $m^*_{\text{CAL}}$ obtained by maximizing agreement between the contrastive metric and human annotations on a calibration set with human AA annotations, and an **example-level threshold** $m^*_{\text{EX}}$ using only within-example scores to avoid the need of calibration data. Following Sarti et al. (2024), we set $m^*_{\text{EX}} = \overline{\mathbf{m}} + \sigma_\mathbf{m}$, where $\overline{\mathbf{m}}$ and $\sigma_\mathbf{m}$ are the average and standard deviation of $\mathbf{m}$ scores for generated tokens.

**CCI Filtering** To extract granular document citations (i.e., colored spans with document indices in Figure 1), we set $s_{\text{CCI}} = a_j^i \geq a^{i*}$, where $a^{i*}$ is either the Top-K or Top-% highest attribution value in $\mathbf{a}^i$, to filter attributed context tokens $c_j \in \text{CCI}(y_i)$. Then, we use the identifier $\text{docid}(c_j)$ of the documents they belong to as citation indices for context-sensitive token $y_i$. Highlights for consecutive tokens citing the same documents are collated into a single span and mapped from subword to word level to facilitate interpretation.

**Sentence-level Aggregation** Following standard sentence-level AA practices, we aggregate token-level citations as the union over all cited documents $\text{docid}(\cdot)$ across context-sensitive tokens in $\mathbf{y}$:

$$\text{MIRAGE}(\mathbf{y}) = \bigcup_{y_i \in \text{CTI}(\mathbf{y})} \text{docid}(c_j) \ \forall c_j \in \text{CCI}(y_i)$$
$$\text{with } s_{\text{CTI}} = m_i \geq m^*, s_{\text{CCI}} = a_j^i \geq a^{i*} \quad (5)$$

In the following sections, we use MIRAGE $_{\text{CAL}}$ and MIRAGE $_{\text{EX}}$ to refer to sentence-level answer attribution using $m^*_{\text{CAL}}$ and $m^*_{\text{EX}}$ thresholds, respectively.

## 4 Agreement with Human Answer Attribution Annotations

We begin our evaluation by comparing MIRAGE predictions to human-produced answer attributions. Importantly, our aim is not to compare several AA approaches to claim optimal faithfulness, but rather evaluate how our proposed framework fares against existing approaches at the task of producing answer attributions from model internals. We employ the XOR-AttriQA dataset (Muller et al., 2023), which, to our knowledge, is the only open dataset with human annotations over RAG outputs produced by a publicly accessible LM.[6] We limit our assessment to open-weights LLMs to ensure that MIRAGE answer attribution can faithfully reflect the model's inner processing towards the natural production of the annotated answer used for evaluation.[7] Moreover, while cross-linguality is not the focus of our work, XOR-AttriQA allows us to assess the robustness of MIRAGE across several languages and its agreement with human annotations compared to an entailment-based system.

### 4.1 Experimental Setup

XOR-AttriQA consists of 500/4720 validation/test tuples, each containing a concise factual query $\mathbf{q}$, a set of retrieved documents that we use as context $\mathbf{c} = \langle \text{doc}_1, \ldots, \text{doc}_k \rangle$, and a single-sentence answer $\mathbf{y}$ produced by an mT5-base model (Xue et al., 2021) fine-tuned on cross-lingual QA in a RAG setup (CORA; Asai et al., 2021).[8] Queries and documents span five languages (Bengali (BN), Finnish (FI), Japanese (JA), Russian (RU), and Telugu (TE)), with no constraint on documents to match the language of the query.[9] Although the RAG generator employs a set of retrieved documents during generation, human annotators were asked to label tuples $(\mathbf{q}, \text{doc}_i, \mathbf{y})$ to indicate whether the information in doc$_i$ supports the generation of $\mathbf{y}$. Importantly, MIRAGE requires extracting model internals in the naturalistic setting that leads to the generation of the desired answer, i.e., the one assessed by human annotators. Hence, we perform a selection procedure to identify XOR-AttriQA examples where the answer produced by filling in the concatenated documents $\mathbf{c}$ in the LM prompt matches the one provided. The resulting subset, which we dub XOR-AttriQA$_{\text{match}}$, contains 142/1144 calibration/test examples and is used for our evaluation.[10]

---

[6]E.g., the human-annotated answers in Bohnet et al. (2022) were generated by PALM 540B (Chowdhery et al., 2023), whose internals are inaccessible.

[7]While these answers could be force-decoded from an open-source model to enable MIRAGE usage, such procedure would likely impact the validity of AA since the selected model would not naturally generate the forced answers.

| Method | Extra Requirements | CCI Filter | BN | FI | JA | RU | TE | Avg. / Std |
|---|---|---|---|---|---|---|---|---|
| NLI$_{\text{ORIG}}$ (Honovich et al.) | 11B NLI model | – | 33.8 | 83.7 | 86.5 | **85.8** | 50.0 | 68.0 / 21.9 |
| NLI$_{\text{MT}}$ (Honovich et al.) | 11B NLI model + MT engine | – | 82.6 | 83.7 | 90.5 | 81.7 | 82.5 | 84.2 / 3.2 |
| MIRAGE$_{\text{CAL}}$ (Ours) | 142 annotated AA examples | Top 3 | 81.7 | **84.2** | 87.8 | 83.3 | 87.0 | 84.8 / 2.3 |
| | | Top 5% | **84.4** | 83.0 | **91.4** | **85.8** | **88.9** | **86.7** / 3.1 |
| MIRAGE$_{\text{EX}}$ (Ours) | – | Top 3 | 80.2 | 78.5 | 83.8 | 77.2 | 75.2 | 79.0 / 2.9 |
| | | Top 5% | <u>81.7</u> | <u>80.1</u> | <u>89.2</u> | <u>84.4</u> | <u>81.8</u> | <u>83.4</u> / 3.2 |

Table 2: Agreement % of MIRAGE and entailment-based baselines with human AA on XOR-AttriQA$_{\text{match}}$ using CORA for RAG. **Extra Requirements**: data/models needed for AA in addition to the RAG model and the current example. **Filter**: $s_{\text{CCI}}$ filtering for saliency scores. **Best overall** and <u>best uncalibrated</u> scores are highlighted.

### 4.2 Entailment-based Baselines

Muller et al. (2023) use an mT5 XXL model fine-tuned on NLI for performing answer attribution on XOR-AttriQA. Since neither the tuned model nor the tuning data are released, we opt to use TRUE[11] (Honovich et al., 2022), a fine-tuned T5 11B model (Raffel et al., 2020), which was shown to highly overlap with human annotation on English answer attribution tasks (Muller et al., 2023; Gao et al., 2023a). We evaluate TRUE agreement with human annotation in two setups. In NLI$_{\text{ORIG}}$, we evaluate the model directly on all examples, including non-English data. While this leads the English-centric TRUE model out-of-distribution, it accounts for real-world scenarios with noisy data, and can be used to assess the robustness of the method in less-resourced settings. Instead, in NLI$_{\text{MT}}$, all queries and documents are machine translated to English using the Google Translate API.[12] While this simplifies the task by ensuring all TRUE inputs are in English, it can lead to information loss caused by imprecise translation.

### 4.3 Results and Analysis

**MIRAGE agrees with human answer attribution** Table 2 presents our results. MIRAGE is found to largely agree with human annotations on XOR-AttriQA$_{\text{match}}$, with scores on par or slightly better than those of the ad-hoc NLI$_{\text{MT}}$ system augmented with automatic translation. Although calibration appears to generally improve MIRAGE's agreement with human annotators, we note that the uncalibrated MIRAGE$_{\text{EX}}$ achieves strong performances despite having no access to external modules or tuning data. These findings confirm that the inner workings of LMs can be used to perform answer attribution, resulting in performances on par with supervised answer attribution approaches even in the absence of annotations for calibration.

**MIRAGE is robust across languages and filtering procedures** Table 2 shows that NLI$_{\text{ORIG}}$ answer attribution performances are largely language-dependent due to the unbalanced multilingual abilities of the TRUE NLI model. This highlights the brittleness of entailment-based approaches in OOD settings, as discussed in Section 2.1. Instead, MIRAGE variants perform similarly across all languages by exploiting the internals of the multilingual RAG model. MIRAGE's performance across languages is comparable to that of NLI$_{\text{MT}}$, which requires an extra translation step to operate on English inputs.

We further validate the robustness of the CCI filtering process by testing percentile values between Top 3-100% for the MIRAGE$_{\text{EX}}$ setting. Figure 4 shows that Top % values between 3 and 20% lead to a comparably high agreement with human annotation, suggesting this filtering threshold can be selected without ad-hoc parameter tuning.

---

[8] https://hf.co/gsarti/cora_mgen
[9] In practice, Muller et al., 2023 report that most retrieved documents are in the same language as the query or in English.
[10] See Appendix A for more details on this selection. Appendix B presents experiments on the full XOR-AttriQA.
[11] https://hf.co/google/t5_xxl_true_nli_mixture
[12] https://cloud.google.com/translate

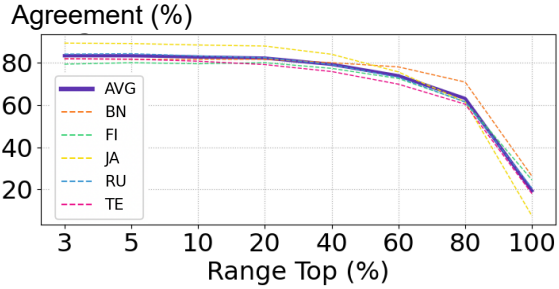

Figure 4: Robustness of MIRAGE$_{EX}$ agreement with human annotations across Top-% CCI filtering thresholds.

## 5 Answer Attribution for Long-form QA

XOR-AttriQA can only provide limited insights for real-world answer attribution evaluation since its examples are sourced from Wikipedia articles, and its answers are very concise. In this section, we extend our evaluation to ELI5 (Fan et al., 2019), a challenging long-form QA dataset that was recently employed to evaluate LLM self-citation capabilities (Gao et al., 2023a). Different from XOR-AttriQA, ELI5 answers are expected to contain multiple sentences of variable length, making it especially fitting to assess MIRAGE context-sensitive token identification capabilities before document attribution. Alongside our quantitative assessment of MIRAGE in relation to self-citation baselines, we conduct a qualitative evaluation of the disagreement between the two methods.

### 5.1 Experimental Setup

**Dataset** The ELI5 dataset contains open-ended why/how/what queries $q$ from the "Explain Like I'm Five" subreddit[13] eliciting long-form multi-sentence answers. For our evaluation, we use the RAG-adapted ELI5 version by Gao et al. (2023a), containing top-5 matching documents $\mathbf{c} = \langle \text{doc}_1, \ldots, \text{doc}_5 \rangle$ retrieved from a filtered version of the Common Crawl (Sphere; Piktus et al., 2021) for every query. The answer attribution task is performed by generating a multi-sentence answer $\mathbf{ans} = \langle \mathbf{y}_1, \ldots, \mathbf{y}_m \rangle$ with an LLM using $(\mathbf{q}, \mathbf{c})$ as inputs, and identifying documents in $\mathbf{c}$ supporting the generation of answer sentence $\mathbf{y}_i, \forall \mathbf{y}_i \in \mathbf{ans}$.

---
[13] https://reddit.com/r/explainlikeimfive

**Models and Answer Attribution Procedure** We select LLaMA 2 7B Chat (Touvron et al., 2023) and Zephyr $\beta$ 7B (Tunstall et al., 2023) for our experiments since they are high-quality open-source LLMs of manageable size. To enable a fair comparison between the tested attribution methods, we first generate answers with inline citations using the self-citation prompt by Gao et al. (2023b).[14] Then, we remove citation tags and use MIRAGE to attribute the resulting answers to retrieved documents. This process ensures that citation quality is compared over the same set of answers, controlling for the variability that could be produced by a different prompt.[15] For more robust results, we perform generation three times using different sampling seeds, and report the averaged scores. Since human-annotated data is not available, we only assess the calibration-free MIRAGE$_{EX}$.

**Entailment-based Evaluation** Differently from the XOR-AttriQA dataset used in Section 4, ELI5 does not contain human annotations of AA. For this reason, and to ensure consistency with Gao et al. (2023a)'s self-citation assessment, we adopt the TRUE model as a high-quality approximation of expected annotation behavior. Despite the potential OOD issues of entailment-based AA highlighted in Section 4, we expect TRUE to perform well on ELI5 since it closely matches the general/scientific knowledge queries in TRUE's fine-tuning corpora and contains only English sentences. To overcome the multi-hop issue when using single documents for entailment-based answer attribution, we follow the ALCE evaluation (Gao et al., 2023a)[16] to measure citation quality as NLI precision and recall (summarized by F1 scores) over the concatenation of retrieved documents.

### 5.2 Results

Results in Table 3 show that MIRAGE provides a significant boost in answer attribution precision and recall for the Zephyr $\beta$ model, while it greatly improves citation recall at the expense of precision for LLaMA 2, resulting in an overall higher F1 score for the MIRAGE$_{EX}$ Top 5% setting. These results confirm that MIRAGE can produce effective answer attributions in longer and more complex

---
[14] The full prompt is provided in Appendix D (Table 9).
[15] For completeness, we also report MIRAGE results without self-citation prompting in Appendix D.
[16] ALCE is an evaluation framework for RAG, evaluating LLM responses in terms of citation quality, correctness, and fluency. More details can be found in Appendix C

| Model | Answer Attrib. | Citation ↑ | | |
|---|---|---|---|---|
| | | Prec. | Rec. | F1 |
| Zephyr $\beta$ | Self-citation | 41.4 | 24.3 | 30.6 |
| | MIRAGE$_{EX}$ Top 3 | 38.3 | 46.2 | 41.9 |
| | MIRAGE$_{EX}$ Top 5% | **44.7** | **46.5** | **45.6** |
| LLaMA 2 | Self-citation | **37.9** | 19.8 | 26.0 |
| | MIRAGE$_{EX}$ Top 3 | 21.8 | **29.6** | 25.1 |
| | MIRAGE$_{EX}$ Top 5% | 26.2 | 29.1 | **27.6** |

Table 3: Answer attribution quality estimated by TRUE for self-citation and MIRAGE on ELI5.

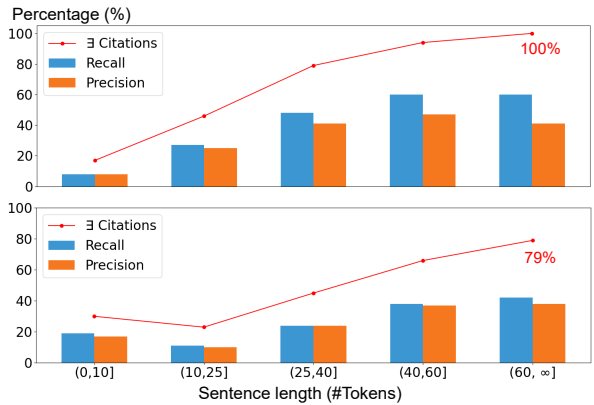

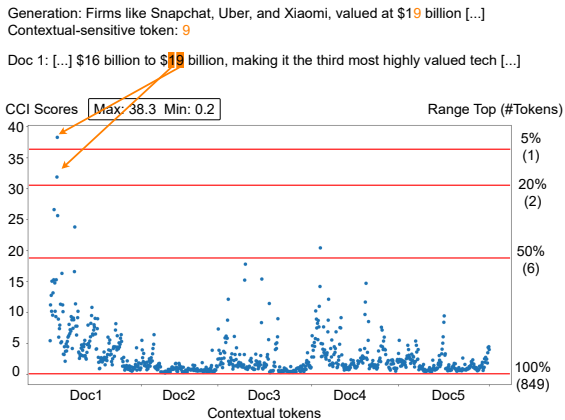

Figure 5: Attribution scores over retrieved documents' tokens for the prediction of context-sensitive token '9'.

settings while employing no external resources like the self-citation approach.

From the comparison between Top 3 and Top 5% CCI filtering strategies, we note that the latter generally results in better performance. This intuitively supports the idea that an adaptive selection strategy is more fitting to accommodate the large variability of attribution scores across different examples. Figure 5 visualizes the distributions of attribution scores $a_j^i$ for an answer produced by Zephyr $\beta$, showing that most context tokens in retrieved documents receive low attribution scores, with only a handful of them contributing to the prediction of the context-sensitive token '9' in the generation. This example also provides an intuitive explanation of the robustness of Top-% selection thresholds discussed in Section 4.3. Ultimately, the Top 5% threshold is sufficient to select the document containing the direct mention of the generated token.

Since the $m^*_{EX}$ threshold used to select context-sensitive tokens by MIRAGE$_{EX}$ depends on the mean and standard deviation of generated answer's scores, we expect that the length of the generated answer might play a role in citation quality. As shown in Figure 6, MIRAGE citation quality is indeed lower for shorter answer sentences. However, a similar trend is observed for self-citation, which is outperformed by MIRAGE for all but the shortest length bin (≤ 10 tokens). The proportion of non-attributed sentences (red line) suggests that the lower quality could be a byproduct of the ALCE evaluation protocol, where non-attributed sentences receive 0 precision/recall. Future availability of human-annotated RAG datasets may shed more light on this effect.

Figure 6: MIRAGE$_{EX}$ (top) and self-citation (bottom) average performance on ELI5 answer sentences binned by length. red: % of sentences with ≥ 1 citation.

### 5.3 Qualitative Analysis of Disagreements

To better understand MIRAGE's performance, we examine some ELI5 examples where MIRAGE disagrees with self-citation on Zephyr $\beta$'s generations. Table 4 and 5 illustrate two cases in which the entailment-based TRUE model results agree with either MIRAGE or self-citation. In Table 4, the answer provided by the model is directly supported by Document [1], as also identified by TRUE. However, self-citation fails to cite the related document at the end of the two sentences. By contrast, MIRAGE attributes several spans to Document [1], resulting in the correct answer attribution for both sentences.

While TRUE achieves high consistency with human judgment (e.g., for the example in Table 4), NLI-based AA can still prove unreliable in cases of high lexical overlap between the answer and supporting documents. Table 5 illustrates one such case, where both self-citation and TRUE attribute the answer to Document [3], whereas MIRAGE does not label any context document as salient for the answer. Here, the answer wrongly states that the bar code can used to ***prevent*** the alarm, while Document [3] mentions that the code can be used to ***cancel*** the alarm after an accidental activation. Thus, despite the high lexical and semantic relatedness, the answer is not supported by Document [3]. The failure of TRUE in this setting highlights the sensitivity of entailment-based systems to surface-level similarity, making them brittle in cases where the model's context usage is not straightforward. Using another sampling seed for the same query produces the answer *"[...] the individual can **cancel** the alarm by providing their password at the keypad"*, which MIRAGE correctly attributes to Document [3].[17]

| INPUT: PROMPT + RETRIEVED DOCS (N=5) + QUERY |
|---|
| **Document [1]** [...] Q. What does it mean for books to be Sensitized or Desensitized? A security strip is embedded into each library book. When a book is checked out, it must be "desensitized" so that it will not set off the alarm when it leaves or enters the library. When the book is returned, it is "sensitized" so that the alarm will be set off should someone attempt to take the book from the library without going through the proper borrowing procedure.<br>**Document [2]** [...] |
| **Query**: How does a small paper library bar code set off the security alarm when you haven't checked a book out? |
| ANSWER ATTRIBUTION RESULTS |
| **Self-citation**: [...] When a book is checked out, it is "desensitized" to prevent the alarm from going off. [∅] When the book is returned, it is "sensitized" so the alarm will sound if the item is taken without authorization. [∅] [...] |
| **MIRAGE**: [...] When a book is checked (1) out, it (1) is "desensitized" (1) to prevent the alarm from going off. [1] When the book (1) is returned, it is "sensitized" (1) so the alarm will sound if the item is taken without authorization. [1] [...] |
| **NLI (TRUE model)**: [1] entails both answer sentences. |

Table 4: Example of self-citation failure using Zephyr β on ELI5. NLI and MIRAGE produce the correct citation, while self-citation does not cite any document ([∅]).

| INPUT: PROMPT + RETRIEVED DOCS (N=5) + QUERY |
|---|
| **Document [2]** [...]<br>**Document [3]** [...] What will happen if you accidentally set off your security system? The siren will sound and it will be loud, but you should be able to stop the siren by entering your code into your keypad. [...]<br>**Document [4]** [...] |
| **Query**: How does a small paper library bar code set off the security alarm when you haven't checked a book out? |
| ANSWER ATTRIBUTION RESULTS |
| **Self-citation**: [...] False alarms can be prevented by entering the code on the keypad, as documented in [3]. [...] |
| **MIRAGE**: [...] False alarms can be prevented by entering the code on the keypad [∅] [...] |
| **NLI (TRUE model)**: [3] entails the answer sentence. |

Table 5: Example showcasing the brittleness of entailment-based AA. MIRAGE correctly finds that the answer cannot be attributed ([∅]), while NLI and self-citation attribute the lexically similar Document [3].

## 6 Conclusion

In this study, we introduced MIRAGE, a novel approach to enhance the faithfulness of answer attribution in RAG systems. By leveraging model internals, MIRAGE effectively addresses the limitations of previous methods based on prompting or external NLI validators. Our experiments demonstrate that MIRAGE produces outputs that strongly agree with human annotations while being more efficient and controllable than its counterparts. Our qualitative analysis shows that MIRAGE can produce faithful attributions that reflect actual context usage during generation, reducing the risk of false positives motivated by surface-level similarity.

In conclusion, MIRAGE represents a promising first step in exploiting interpretability insights to develop faithful answer attribution methods, paving the way for the usage of LLM-powered question-answering systems in mission-critical applications.

## Limitations

**LLMs Optimized for Self-citation** In this study, we focus our analysis on models that are not explicitly trained to perform self-citation and can provide citations only when prompted to do so. While recent systems include self-citation in their optimization scheme for RAG applications[18], we believe incorporating model internals in the attribution process will remain a valuable and inexpensive method to ensure faithful answer attributions.

**Brittleness of NLI-based Evaluation** Following Gao et al. (2023a), the evaluation of Section 5 employs the NLI-based system TRUE due to the lack of AA-annotated answers produced by open-source LLMs. However, using the predictions of NLI models as AA references is far from ideal in light of their brittleness in challenging scenarios and their tendency to exploit shallow heuristics. While the ELI5 dataset is reasonably in-domain for the TRUE model, this factor might still undermine the reliability of some of our quantitative evalua-tion results. Future work should produce a wider variety of annotated datasets for reproducible answer attribution using open-source LLMs, enabling us to extend our analysis to a broader set of languages and model sizes and ultimately enhance the robustness of our findings.

---
[17]This and other examples are provided in Appendix E.

[18]For example, the Command-R models: https://huggingface.co/CohereForAI/c4ai-command-r-plus

**Applicability to Other Domains and Models** Our evaluation is conducted on relatively homogeneous QA datasets and does not include language models with >7B parameters. This limits the generalizability of our findings to other domains and larger models. Future work should extend our analysis to a broader range of domains and model sizes to further validate the robustness and applicability of MIRAGE. This said, we expect MIRAGE to be less vulnerable to language and quality shifts compared to existing AA methods that depend on external validators or on the model's instruction-following abilities.

**Scalability of MIRAGE on Longer Context** The computational cost for the simple gradient-based version of MIRAGE proposed in this work is $2O(F) + |\text{CTI}(\mathbf{y})| \cdot O(B)$, where $O(F), O(B)$ are respectively the costs of a forward and a backward pass with the LLM, and $|\text{CTI}(\mathbf{y})|$ is the number of tokens selected by the CTI step. While CTI effectively limits the expensive backward component in the MIRAGE computation, its cost is bound to increase significantly for larger models and context sizes. When applying MIRAGE to LLMs with <10B parameters, we note that its cost can be comparable or lower to supervised models like TRUE, requiring several forward passes using a large 11B LLM. Importantly, MIRAGE is a flexible framework that can be implemented using different feature attribution methods in the CCI step, including lightweight techniques requiring only forward passes (e.g., Attention Rollout (Abnar and Zuidema, 2020), Value Zeroing (Mohebbi et al., 2023), or ALTI-Logit (Ferrando et al., 2023). Finally, a promising perspective for scaling to larger LLMs could be to assess whether MIRAGE-produced AAs remain accurate when force-decoding the original model's answer from a different LLM with fewer parameters.

**MIRAGE's Parametrization and Choice of Attribution Method** While Section 4.1 highlights the robustness of MIRAGE to various CCI filtering thresholds, the method still requires non-trivial parametrization. In particular, we emphasize that the choice of the attribution method employed to generate attribution scores in the CCI step can significantly impact the faithfulness of the resulting answer attributions. Although we used a relatively simple gradient-based approach in this study, our proposed framework is method-agnostic. We leave the evaluation of modern feature attribution techniques, such as the ones mentioned in the previous paragraph, to future work to further improve MIRAGE applicability in real-world settings.

## Acknowledgments

The authors have received funding from the Dutch Research Council (NWO): JQ is supported by NWA-ORC project LESSEN (grant nr. NWA.1389.20.183), GS is supported by NWA-ORC project InDeep (NWA.1292.19.399), AB is supported by the above as well as NWO Talent Programme (VI.Vidi.221C.009). RF is supported by the European Research Council (ERC) under European Union's Horizon 2020 programme (No. 819455).

## A Construction of XOR-AttriQA$_{\text{match}}$

XOR-AttriQA$_{\text{match}}$ is a subset of the original XOR-AttriQA containing only examples for which our LLM generation matches exactly the answer annotated in the dataset. Replicating the original answer generation process is challenging since the original ordering of the documents doc$_i$ in **c** is unavailable.[19] To maximize the chances of replication, we attempt to restore the original document sequence by randomly shuffling the order of doc$_i$s until LLM can naturally predict the answer **y** (otherwise, at most 200 iterations), as shown in Algorithm 1. The statistics of the original XOR-AttriQA and XOR-AttriQA$_{\text{match}}$ are shown in Table 7.

---

[19] Muller et al. 2023 only provide the split documents without the original ordering.

| Method | Extra Requirements | BN | FI | JA | RU | TE | Avg. / Std |
|---|---|---|---|---|---|---|---|
| mT5 XXL $_{\text{NLI}}$ (Honovich et al.) | 11B NLI model (250 FT ex.) | 81.9 | 80.9 | 94.5 | 87.1 | 88.7 | 86.6 / 4.9 |
| | 11B NLI model (100k FT ex.) | 89.4 | 88.3 | 91.5 | 91.0 | 92.4 | 90.5 / 1.5 |
| | 11B NLI model (1M FT ex.) | 91.1 | 90.4 | 93.0 | 92.9 | 93.8 | 92.2 / 1.3 |
| PALM2$_{\text{LORA}}$ (Anil et al.) | 540B LLM (250 FT ex.) | 91.5 | 88.3 | 94.7 | 93.7 | 93.7 | 92.4 / 2.3 |
| PALM2 (Anil et al.) | 540B LLM (250 FT ex.) | **92.3** | **92.6** | **96.4** | **94.5** | **94.8** | **94.1** / 1.5 |
| PALM2 (Anil et al.) | 540B LLM (4-shot prompting) | 91.5 | 87.4 | 92.0 | 90.5 | 90.6 | 90.4 / 1.6 |
| PALM2$_{\text{CoT}}$ (Anil et al.) | 540B LLM (4-shot prompting) | 83.7 | 78.8 | 71.7 | 81.9 | 84.7 | 80.2 / 4.7 |
| MIRAGE$_{\text{CAL}}$ (Ours) | 500 AA calibration ex. | <u>82.2</u> | <u>82.5</u> | <u>92.0</u> | <u>87.7</u> | <u>90.2</u> | <u>86.9</u> / 4.0 |
| MIRAGE$_{\text{EX}}$ (Ours) | – | 79.0 | 74.1 | 90.8 | 82.6 | 86.9 | 82.7 / 5.8 |

Table 6: Agreement % of MIRAGE and entailment-based baselines with human AA on the full XOR-AttriQA using CORA for RAG (annotated answers not matching the LM's natural generation are force-decoded). **Extra Requirements**: data/models needed for AA in addition to the RAG model itself. **Best overall** and <u>best validator-free</u> scores are highlighted. PALM and mT5 results are taken from Muller et al. (2023).

---

**Algorithm 1** Restore original document sequence

**Input:** $\{Doc_1, ..., Doc_n\}, query, answer, \mathbb{M}$
1: $iter \leftarrow 0, \ found = False$
2: **while** $iter < 200$ **do**
3:    $pred \leftarrow \mathbb{M}(\{Doc_1, ..., Doc_n\}, query)$
4:    **if** $pred == answer$ **then**
5:       $found = True$ **break**
6:    **else**
7:       Shuffle($\{Doc_1, ..., Doc_n\}$)
8:    **end if**
9:    $iter \mathrel{+}= 1$
10: **end while**
11: **if** $found$ **then**
12:    **return** $\{Doc_1, ..., Doc_n\}$
13: **end if**

| Dataset | BN | FI | JA | RU | TE | Total |
|---|---|---|---|---|---|---|
| Orig. | 1407 | 659 | 1066 | 954 | 634 | 4720 |
| Match | 274 | 214 | 232 | 254 | 170 | 1144 |

Table 7: Statistic for test sets of the original (Orig.) XOR-AttriQA and XOR-AttriQA$_{\text{match}}$.

## B Answer Attribution on the Full XOR-AttriQA

Differently from the concatenation setup in Section 4.1, we also test MIRAGE on the full XOR-AttriQA dataset by constraining CORA generation to match the annotated answer **y**. We adopt a procedure similar to Muller et al. (2023) by considering a single document-answer pair $(doc_i, \mathbf{y})$ at a time, and using MIRAGE's CTI step to detect whether **y** is sensitive to the context $doc_i$. Results in Table 6 show that MIRAGE achieves performances in line with other AA methods despite these approaches employing ad-hoc validators trained with as many as 540B parameters.

## C ALCE Evaluation Benchmark

Gao et al. (2023a) propose ALCE, an evaluation framework for RAG QA tasks. ALCE assesses the LLMs' response from three diverse aspects: citation quality, correctness, and fluency. **Citation quality** evaluates the answer attribution performance with recall and precision scores. The <u>recall</u> score calculates if the concatenation of the cited documents entails the generated sentence. The <u>precision</u> measures if each document is cited precisely by verifying if the concatenated text still entails the generation whenever one of the documents is removed. We further calculate F1 scores to summarize the overall performance. **Correctness** checks whether the generated answer entails the golden reference answer according to the NLI model TRUE. Gold-reference answers are provided in the original dataset, and some were summarized by Gao et al. (2023b) by using GPT-4 in case they were too long. **Fluency** reflects the coherence and fluency of the generated response according to MAUVE (Pillutla et al., 2021), a popular NLG metric. We report the average score for all instances for each evaluation metric.

## D ELI5 Evaluation with Standard Prompt

In the main experiments, we use self-citation prompts by Gao et al. (2023a) for MIRAGE answer attribution to control for the effect of different prompts on model responses, enabling a direct comparison with self-citation. In Table 8, we provide additional results where a standard prompt without citation instructions is used ("Standard" prompt in Table 9). We observe the overall citation quality of MIRAGE drops when a standard prompt is used instead of self-citation instructions. We conjecture this might be due to answers that are, in general, less attributable to the provided context due to a lack of explicit instructions to do so. We also observe higher correctness and fluency in the standard prompt setting, suggesting a trade-off between answer and citation quality.

| Model | Prompt | Answer Attribution | Filter | Citation↑ | | | Correctness↑ | Fluency↑ |
|---|---|---|---|---|---|---|---|---|
| | | | | Prec. | Rec. | F1 | | |
| Zephyr | Self-citation | Self-citation | - | 41.4 | 24.3 | 30.6 | 9.9 | 28.6 |
| | | MIRAGE_EX | Top 3 | 38.3 | 46.2 | 41.9 | | |
| | | | Top 5% | **44.7** | **46.5** | **45.6** | | |
| | Standard | MIRAGE_EX | Top 3 | 29.8 | 34.5 | 32.0 | **11.3** | **34.3** |
| | | | Top 5% | 34.1 | 34.2 | 34.1 | | |
| LLaMA | Self-citation | Self-citation | - | **37.9** | 19.8 | 26.0 | 11.8 | 24.8 |
| | | MIRAGE_EX | Top 3 | 21.8 | **29.6** | 25.1 | | |
| | | | Top 5% | 26.2 | 29.1 | **27.6** | | |
| | Standard | MIRAGE_EX | Top 3 | 17.8 | 20.9 | 19.2 | **13.0** | **26.4** |
| | | | Top 5% | 21.1 | 20.1 | 20.6 | | |

Table 8: Citation quality (F1, Recall, Precision), correctness, and fluency of self-citation and MIRAGE on ELI5 when using self-citation instruction and standard prompts of Table 9.

| SELF-CITATION (Gao et al., 2023a) |
|---|
| Instruction: Write an accurate, engaging, and concise answer for the given question using only the provided search results (some of which might be irrelevant) and cite them properly. Use an unbiased and journalistic tone. Always cite for any factual claim. When citing several search results, use [1][2][3]. Cite at least one document and at most three documents in each sentence. If multiple documents support the sentence, only cite a minimum sufficient subset of the documents. |
| STANDARD |
| Instruction: Write an accurate, engaging, and concise answer for the given question using only the provided search results (some of which might be irrelevant). Use an unbiased and journalistic tone. |

Table 9: Self-citation prompt, taken from Gao et al., 2023a, and standard prompt with no citation instruction.

# E More Examples of Disagreement

Table 10 to 12 show three cases where MIRAGE answer attributions disagree with self-citation attributions of the same generation[20]. We adopt the Top-5% threshold for CCI Filtering. In Table 10, the generated answer becomes the consistent description '*cancel* the alarm' as mentioned in Document [3]. In this case, MIRAGE attributes this sentence to the corresponding Document [3] while NLI maintains its attribution of Document [3] due to lexical overlap, as suggested in Section 5.3.

---
[20]Note that we remove citation tags in self-citation generated answers and use MIRAGE to attribute the resulting answers, as introduced in Section 5

| INPUT: PROMPT + RETRIEVED DOCS (N=5) + QUERY |
|---|
| **Document [2]** [...] |
| **Document [3]** [...] What will happen if you accidentally set off your security system? The siren will sound and it will be loud, but you should be able to stop the siren by entering your code into your keypad. [...] |
| **Document [4]** [...] |
| **Query**: How does a small paper library bar code set off the security alarm when you haven't checked a book out. |
| ANSWER ATTRIBUTION RESULTS |
| **Self-citation**: [...] The response center may contact the individual to confirm if it's a false alarm, and the individual can cancel the alarm by providing their password at the keypad. [∅] [...] |
| **MIRAGE**: ... The response center(3) may(3) contact the individual to confirm if it(3)'s a false alarm, and the individual(3) can cancel(3) the alarm by providing(3) their password at the keypad(3).[3] ... |
| **NLI (TRUE model)**: [3] entails the answer sentence. |

Table 10: Example described in Section 5.3: MIRAGE attributes the generation to Document [3] when *cancel* is used instead of *prevent* (Table 5).

On several occasions, we observe that MIRAGE attributes all occurrences of lexically similar tokens in the context when the LLM is generating the same word. For example, in Table 11 the named entity "Science ABC" is mentioned in both Document [1] and [4], and MIRAGE finds both occurrences as salient towards the prediction of the same entity in the output. Similarly, in Table 12, the generated word 'Document' is attributed to the previous mentions of the same word in the context. In both cases, when moving from token-level to sentence-level AA, this dependence would result in wrong AA according to NLI, since the documents are not entailing the answer, but rather making a specific token more likely. These cases reflect the possible discrepancy between AA intended as logical entailment and actual context usage during generation. Future work could explore more elaborate ways to aggregate granular information at sentence level while preserving faithfulness to context usage.

| INPUT: PROMPT + RETRIEVED DOCS (N=5) + QUERY | INPUT: PROMPT + RETRIEVED DOCS (N=5) + QUERY |
|---|---|
| **Document [1]**(Title: Why Do Airlines Sell More Tickets Than The Aircraft's Capacity? » Science ABC): [...] They know that some people simply won't show up for their appointments [...] To account for this discrepancy, where the flight would have to fly with some very expensive seats empty, they decide to opt for a more profitable strategy. Assuming that out of the maximum capacity of 200 seats only 185 show up for the flight, this leaves 15 seats available. The airline could try to sell these seats quickly to avoid losses, but this rarely happens. What the airline does instead is try to sell 220 seats in advance. **Document [2]** [...] **Document [3]** [...] **Document [4]**(Title: Why Do Airlines Sell More Tickets Than The Aircraft's Capacity? » Science ABC): your own interests, as the airlines have clearly looked after theirs. First of all, if the airline is offering you a seat on a different flight, check if that seat is definitely available. If it is also full and it looks like you might be bumped yet again, you might as well be stranded! If the airline is offering compensation, such as free meals, hotel costs, and transportation between the airport and the hotel, look for the most profitable outcome. The airline might offer you vouchers for future flights, but since airline employees are given guidelines for negotiation, you might **Document [5]** [...] | **Document [1]**(Title: Why Do Airlines Sell More Tickets Than The Aircraft's Capacity? » Science ABC): board the aircraft. They know that some people simply won't show up for their appointments (just like at hospitals, hotels, restaurants etc.). To account for this discrepancy, where the flight would have to fly with some very expensive seats empty, they decide to opt for a more profitable strategy. Assuming that out of the maximum capacity of 200 seats only 185 show up for the flight, this leaves 15 seats available. The airline could try to sell these seats quickly to avoid losses, but this rarely happens. What the airline does instead is try to sell 220 seats in advance. **Document [2]** [...] **Document [3]** [...] **Document [4]**(Title: Why Do Airlines Sell More Tickets Than The Aircraft's Capacity? » Science ABC): your own interests, as the airlines have clearly looked after theirs. First of all, if the airline is offering you a seat on a different flight, check if that seat is definitely available. If it is also full and it looks like you might be bumped yet again, you might as well be stranded! If the airline is offering compensation, such as free meals, hotel costs, and transportation between the airport and the hotel, look for the most profitable outcome. The airline might offer you vouchers for future flights, but since airline employees are given guidelines for negotiation, you might **Document [5]** [...] |
| **Query**: Why are airlines able to sell more seats than are available on a flight? | **Query**: Why are airlines able to sell more seats than are available on a flight? |
| ANSWER ATTRIBUTION RESULTS | ANSWER ATTRIBUTION RESULTS |
| **Self-citation**: [...] According to a report from Science ABC [1], airlines assume that out of the maximum capacity of 200 seats, only 185 will actually board the aircraft. To avoid empty seats and potential losses, airlines opt to sell up to 220 seats in advance [1]. [...] | **Self-citation**: [...] According to a source in Document [1], airlines assume that not all passengers will show up for their flights, leaving some seats empty. [...] |
| **MIRAGE**: [...] According to a report from Science ABC $^{(1)(4)}$, airlines assume that out $^{(1)}$ of the maximum capacity $^{(1)}$ of 200 seats $^{(1)}$, only 185 $^{(1)}$ will actually board the aircraft. [1][4] [...] | **MIRAGE**: [...] According to a source in Document $^{(4)}$, airlines assume that not $^{(1)}$ all passengers will show up for their flights, leaving some seats empty. [1][4] [...] |
| **NLI (TRUE model)**: Only [1] entails the answer sentence. | **NLI (TRUE model)**: Only [1] entails the answer sentence. |

Table 11: Example of counterintuitive MIRAGE attribution: the generation is grounded to Document [4] because the named entity 'Science ABC' is made more probable by the occurrence of the same named entity 'Science ABC'. However, Document [4] does not logically entail the answer.

Table 12: Example of counterintuitive MIRAGE attribution: Document [4] is attributed by MIRAGE due to the repetition of the keyword 'Document'.